\documentclass[sigconf,authorversion,screen,balance]{acmart}

\AtBeginDocument{%
  }

\copyrightyear{2022}
\acmYear{2022}
\setcopyright{acmcopyright}
\acmConference[DigiPro '22]{The Digital Production Symposium}{August 7, 2022}{Vancouver, BC, Canada}
\acmBooktitle{The Digital Production Symposium (DigiPro '22), August 7, 2022, Vancouver, BC, Canada}
\acmPrice{15.00}
\acmDOI{10.1145/3543664.3543672}
\acmISBN{978-1-4503-9418-5/22/08}

\usepackage{xspace}
\usepackage{microtype}
\definecolor{purple}{rgb}{0.3, 0, 0.6}
\definecolor{darkred}{rgb}{0.6, 0.1, 0.1}

\newcommand{\FIXLATER}[1]{}  % fix this after submitting thesis
\newcommand{\LONGVERSION}[1]{}  % for longer version, e.g. thesis or journal

\newcommand{\Secref}[1]{{Section~\ref{#1}}}   % spelling of "section" reference
\newcommand{\Figref}[1]{{Fig.~\ref{#1}}}   % spelling of figure reference

\newcommand{\link}[1]{\href{#1}{[link]}}

\newcommand{\dd}{\mathbf{d}}

\newcommand{\mm}{\mathbf{m}}

\newcommand{\qq}{\mathbf{q}}

\newcommand{\ww}{\mathbf{w}}
\newcommand{\xx}{\mathbf{x}}

\newcommand{\norm}[1]{\left\lVert#1\right\rVert}  % better looking norm, I think it scales vertically if the thing inside is tall

\graphicspath{{assets/}}     % organize your images and other figures under media/ folder

\usepackage{cuted} % for secondary title for supplementary
\begin{document}

\title{FDLS: A Deep Learning Approach to Production Quality, Controllable, and Retargetable Facial Performances}

\author{Wan-Duo Kurt Ma}
\orcid{1234-5678-9012-3456}
\affiliation{
 \institution{Weta Digital + Unity}
 %\streetaddress{104 Jamestown Rd}
 \city{Wellington}
 \country{New Zealand}}
\email{kma@wetafx.co.nz}

\author{Muhammad Ghifary}
\authornote{currently with PT.~Bank Rakyat Indonesia (Persero), Tbk.}
\orcid{1234-5678-9012}
%\email{webmaster@marysville-ohio.com}
\affiliation{%
  \institution{Weta Digital}
  %\streetaddress{P.O. Box 1212}
  \city{Wellington}
  \country{New Zealand}
}
\email{mghifary@gmail.com}

\author{J.P.~Lewis}
\authornote{currently with NVIDIA Research}
\affiliation{%
  \institution{Weta Digital}
  %\streetaddress{1 Th{\o}rv{\"a}ld Circle}
  \city{Wellington}
  \country{New Zealand}}
\email{noisebrain@gmail.com}

\author{Byungkuk Choi}
\affiliation{%
  \institution{Weta Digital + Unity}
  % \streetaddress{8600 Datapoint Drive}
  \city{Seoul}
  % \state{Texas}
  \country{Korea}
  \postcode{78229}}
\email{bchoi@wetafx.co.nz}

\author{Haekwang Eom}
\affiliation{%
  \institution{Weta Digital + Unity}
  % \streetaddress{1 Th{\o}rv{\"a}ld Circle}
  \city{Seoul}
  \country{Korea}}
\email{heom@wetafx.co.nz}

\renewcommand{\shortauthors}{Ma et al.}

\begin{abstract}
Visual effects commonly requires both the creation of realistic synthetic humans as well as retargeting actors' performances to  humanoid characters such as aliens and monsters. Achieving the expressive performances demanded in entertainment requires manipulating complex models with hundreds of parameters.
Full creative control requires the freedom to make edits at any stage of the production, which prohibits the use of a fully automatic ``black box'' solution with uninterpretable parameters. On the other hand, producing realistic animation with these sophisticated models is difficult and laborious.
~\\
This paper describes FDLS (Facial Deep Learning Solver \footnote{\href{https://www.youtube.com/watch?v=39W4eGjMFiA}{https://www.youtube.com/watch?v=39W4eGjMFiA}}), which is Weta Digital's solution to these challenges. FDLS adopts a coarse-to-fine and \emph{human-in-the-loop} strategy, allowing a solved performance to be verified and (if needed) edited at several stages in the solving process. % with differing levels of granularity
To train FDLS, we first transform the raw motion-captured data into robust graph features. The feature extraction algorithms were devised after carefully observing the artists' interpretation of the 3d facial landmarks.
 Secondly, based on the observation that the artists typically finalize the jaw pass animation before proceeding to finer detail, we solve for the jaw motion first and predict  fine expressions with region-based networks conditioned on the jaw position. Finally, artists can optionally invoke a non-linear finetuning process on top of the FDLS solution to follow the motion-captured virtual markers as closely as possible. FDLS supports editing if needed to improve the results of the deep learning solution and it can handle small daily changes in the actor's face shape.
 ~\\
 FDLS permits reliable and production-quality performance solving with minimal training and little or no manual effort in many cases, while also allowing the solve to be guided and edited in unusual and difficult cases. The system has been under development for several years and has been used in major movies.
\end{abstract}


\ccsdesc[500]{Computing methodologies~Animation}
%\ccsdesc[500]{Computer systems organization~Embedded systems}

\keywords{Facial animation, Deep learning, Motion capture, Optimization}

\maketitle

\section{Introduction}

% single image
\begin{figure}[H]
    \centering
    \includegraphics[width=.5\textwidth]{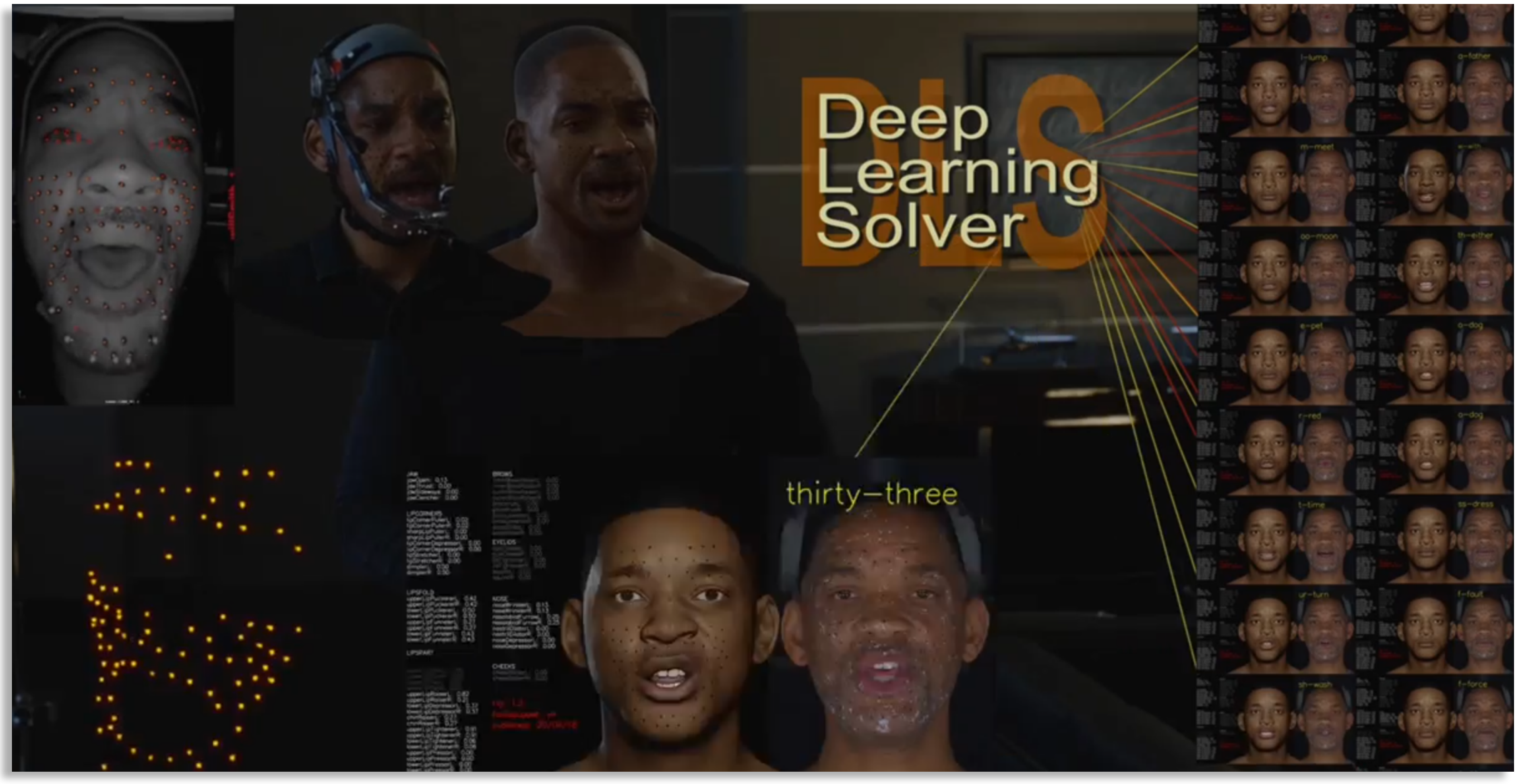}
    \caption{Facial Deep Learning Solve (FDLS) is an animator oriented tool system for solving facial animation given motion captured sparse marker set and limited training data.
    }

    \label{img}
\end{figure}

Digital characters representing either humans or creatures are ubiquitous in the film and video game industries. Creating believable performances with these characters is one of the grand challenges of computer graphics.
% jp wording to restrict the challenge to just the "animation", excluding modeling, rendering which are not addressed by this paper
%One of the grand challenges is to model and animate high fidelity facial expression ...
%
%Modeling and animating high fidelity digital characters are some of the grand challenges in the visual effect industry.
%
%Creating high fidelity human face animation is one of the grand challenges in computer graphics.
Realistic digital characters are often animated using performances captured from actors, as this allows the characters to inherit the ``personality'' of selected actors.
On the other hand, performance capture presents its own difficulties arising from the complexity of the models, the limited amounts of data available (to retain fidelity to a particular actor's performance of a particular character, it may not be possible to use data from other actors, nor even from the same actor performing a different character), and the need to potentially edit the resulting performances.

This paper describes FDLS, Weta Digital's solution to this challenge.
FDLS has been in development since 2016 and has been used on movies such as \emph{Gemini Man} \footnote{Gemini Man VFX Breakdown - Junior | Weta Digital: ``All of this was to ensure we captured the youthfulness, not just the likeness, of a 23-year-old Will Smith.'' \href{https://www.youtube.com/watch?v=V1lTuNLuIO8}{https://www.youtube.com/watch?v=V1lTuNLuIO8}}. Our system has the following assumptions and requirements:

\begin{itemize}

\item \textbf{Production quality,} with high and verifiable accuracy. For example we assume that an actor's face shape may change slightly across days, and account for this. A stereo or mono head-mounted camera (HMC) with virtual (painted) markers is used in order to allow simultaneous capture of the head and body motion, since marker-free tracking methods are considered to not yet be sufficiently accurate.

\item \textbf{Perceptual evaluation} is used
(as is commonly the case in image quality assessment as well \cite{zhang2018perceptual}), with guidance provided by the VFX supervisor.

% submission version
%\item \textbf{Retargetable and editable results.} The actor's performance often needs to be re-targeted to other characters, including humanoid aliens and monsters. Editing is needed both to retain full creative control, and in infrequent cases where the actor's expression is not conveyed correctly on a target character with very different proportions (e.g.~a dragon). Thus FDLS excels at post editing on rig (aka~\emph{puppet}) parameters such as FACS (Facial Action Coding System) action units \cite{ekman1978facial}.
%jp not sure if the "puppet" word is used outside weta, maybe it is. kma: I see some paper uses puppet. jp- ok!

\item \textbf{Editable and retargetable results.} Editing is needed to retain full creative control. A closed ``black box'' solver is not suitable
%, regardless of accuracy,
because the supervisor may \emph{require} that the performance be edited for various reasons. The actor's performance often needs to be re-targeted to other characters, such as humanoid aliens and monsters. Occasionally a particular expression may not be conveyed correctly on a character with very different proportions (e.g.~a dragon). FDLS excels at post editing on rig (aka~\emph{puppet}) parameters such as FACS (Facial Action Coding System) action units \cite{ekman1978facial}.

\item We target \textbf{expert animators} rather than casual or novice users, while greatly reducing the effort needed to manipulate complex facial models,
\item The ability to handle a rig with \textbf{nonlinear effects,} including intermediate and corrective blendshapes \cite{osipa10} and additional deformers e.g.~for handling collisions between the lips.

\item As mentioned above, we assume \textbf{limited training data} relative to the massive datasets commonly used to train deep neural networks.
To address this issue we introduce strong inductive biases by way of specifically designed vertex graph features, that better capture the facial expression while remaining insensitive to irrelevant factors. %including the relative position of camera.
% allows to bootstrap

\item We require performance capture that has been \textbf{stabilized} to remove rigid motion of the skull \cite{derekbeelerStabilization,LamarreLD18}. We use a stereo or mono head-mounted camera, however the use of a head-mounted camera is not itself sufficient for stabilization since the camera may slip or the actor's scalp may move with respect to the skull.

\end{itemize}

This paper describes the "face solving" component of an overall character pipeline. Other components such as modeling, rigging, adaptation of the rig to specific actors \cite{ma-blendshape-16,seolactorspecific}, tracking, stabilization, muscle simulation, and rendering are separate topics not covered here.

The primary contributions of this paper are:
\begin{itemize}
    \item A presentation of the motivation and design decisions for Weta Digital's Facial Deep Learning Solver (FDLS) system, a novel \emph{multi-stage face solving} approach for production quality performance-driven facial animation using deep neural networks.
    \item Graph features that allow accurate animation solving with limited training data.

    \item A nonlinear post-finetuning method conditioned on jaw animation to allow better matching of some difficult face expressions.
    \item A method to accommodate the changes in the marker layout or actor's face shape without requiring re-training of the network.
\end{itemize}

\section{Related work}

% todo: zhang spacetime faces, li/bregler,
The literature on facial performance capture dates from \cite{Williams90} and is too large to fully review, for example \cite{zollhofer-monocularfacetrackSTAR-18} primarily targets the subset of monocular and optimization-based approaches yet contains more than 200 references.  We will focus on methods that using deep learning as well as those that give professional quality and editable results, while providing a few pointers to other literature. %that produces non-editable representations such as dense meshes or uninterpretable (e.g.~PCA) parameters.
Broader surveys of facial performance capture and related areas include \cite{zollhofer-monocularfacetrackSTAR-18,klehm-survey-15,tagliasacchi-nonrigid-16}.

%While the blendshape models commonly used in computer graphics entertainment applications are distinguished by their artist-friendly interpretable parameters,
Computer vision research often targets fully automatic approaches that produce non-editable representations such as dense meshes or uninterpretable (e.g.~PCA) parameters. \cite{BV99} introduced the morphable model approach in which scanned facial geometry and texture are approximated with PCA linear subspaces.
This approach has been extended in a large body of research over the past two decades \cite{egger-3dmmsurvey-20}.
% jp- may be correct, but not sure about tena, safer to cite survey for now
% revisit for final
%in various ways to achieve performance tracking in real-time \cite{FLAME17,Tena11}.
Performance capture of dense facial meshes usually requires complex camera setups \cite{Fyffe14,Bhat13,Bradley10} and multi-view stereo or photogrammetric reconstruction \cite{Fyffe14,furukawa-densemocap-2009,Zhang04}. Other research introduces algorithms for specific facial regions such as the lips \cite{Garrido16} and eyelids \cite{Bermano15}.

Real-time tracking and re-targeting from monocular RGB video has seen widespread use in face ``filters'' for video conferencing systems. While these methods have impressive ease of use and robustness, they do not offer \emph{verifiable} accuracy and may not capture fine facial detail, as well as  being prone to errors from significant lighting changes and occlusions. See \cite{zollhofer-monocularfacetrackSTAR-18} for a recent survey of the literature in this area.
Higher quality markerless capture has also been demonstrated recently \cite{disneyAnyma}, albeit  under constrained viewing and lighting conditions that would prevent simultaneous capture of facial and body performances. Though there are differences in opinion, some believe that such simultaneous capture is essential to capture natural correlations between face and body motion.

The facial animation solvers used in visual effects production and games often use physical or virtual (painted) markers as input, due to accuracy requirements and reliability of marker-based systems \cite{Bickel07, Huang11}.
% not necessary: The solver must then fit the facial rig to these motion captured landmarks.
% Marker-driven facial animation solver is generally the task of fitting the facial rig to the motion captured landmarks \cite{Zhang04}. %jp Zhang was not marker based I think, moved it to multiview stereo section
%Markers is often preferred in visual effects production because of its reliability \cite{Bickel07, Huang11} at the expense human efforts.
However, these approaches have less ability to capture fine detail in complex regions of the face, in particular around the lips. One of the accommodations in these marker-driven methods is to add additional information extracted from the captured footage. For instance, \cite{Bhat13} adds separately tracked contours
% manually traces the silhouette contour
% jp paper says "tracked contours", probably manual but doesn't exactly say
of the lips and eyelid to the solver's input.

% \cite{seol-spacetimeblendshape-2012} % jp this paper is more relevant, is about editing performances whereas the wpsd paper is mainly a hybrid linear/nonlinear (pose space deformation) rig

% Leave this out:
% \cite{roberts-simplify-19} introduced an algorithm to optimally simplify motion capture or dense solved performances into sparse keyframes to support editing.

Keyframe animation is widely used as it allows artists to specify intuitively understandable parameters at a subset of frames, and then interpolate these parameters to the remaining frames using (for example) Catmull-Rom \cite{LD08} or B-spline \cite{Choe01} curves.
%Keyframing-based solution is the intuitive and widely used by the artists who specify the a set of specific parameters (e.g., ~blendshape weights) at particular frame. Then, the final animation of the clip is constructed by the interpolation of the keys either linearly, or to fit Catmull-Rom spline \cite{LD08} or B-spline curve \cite{Choe01}.
On the other hand, keyframe animation is quite laborious, and
% One industry professional characterized it as requiring on the ballpark of one day of effort for each second of finished animation \cite{X}.
various techniques have been proposed to speed up the artist's workflow. For example \cite{seolretargeting11} introduced a successive refinement scheme in which large motions (e.g.~the jaw) are solved before the fine details. Temporal animation editing has been developed in several directions, includings gradient-domain space-time editing\cite{seol-spacetimeblendshape-2012} and systems that allow interactive tuning of the solution using nonlinear regression with radial basis function networks \cite{SeolC19, Seol14}.
%\cite{seol-spacetimeblendshape-2012} introduce a gradient-domain space-time editing method.
%solves the motion re-targeting problem with movement matching.
Intuitive user guidance in the form of 2D sketches has been used for generating plausible lip corrections \cite{Dinev2018UserGuidedLC}, guiding blendshape models \cite{cetinaslan-sketchblendshape} and doing space-time editing of general animation \cite{choi-sketchimo-16}. \cite{berson-rnnedit-2020, berson-faceedit-2019} use a learning based approach to perform temporal animation editing.

% jp - vote that it is not necessary to say this here, say it in 3.6, and it is also mentioned in the contributions in the introduction, as well as a sentence in the conclusion (the commented paragraph below)
% Similar to the temporal editing, FDLS uses anchor pose that allows the inconsistency of the facial basis and the unseen motion captured markers.

% trying new order: 1) computer vision automatic uninterpretable, 2) blendshape/parametric/editable 3) CNN/deep learning based

% \cite{laine-performancecapture-17, Klaudiny17} multiview in training, monocular evaluation,
% Laine assumes static cameras, convolutional helps with lack of stabilization, regresses mesh as output

Performance capture systems developed in production settings
target high quality and (in most cases) editable rig representations.
% share similar concerns and tradeoffs with our work.
\cite{smith-emotionchallenge-17} demonstrates high-quality blendshape solving using an optimization approach.  Convolutional neural networks have been employed to regress directly from video to dense meshes \cite{laine-performancecapture-17}. \cite{moser-DD-18} describes a production-proven system for regressing 3D marker positions.
In addition to this research, there are  several commercial systems that address aspects of professional performance capture, including DI4D, Dynamixyz, Synthesia, Medusa, and Imagemetrics \cite{software}.

``Deepfake'' systems have recently been applied to face replacement for stunt doubles \cite{wetaFFS}. In these systems an autoencoder with separate decoders for the actor and stunt double is used to learn a shared latent space, allowing the double's performance to be decoded using the actor's likeness.
In this stunt double application the resolution limitations of current neural rendering is not a limiting issue, since the view of the stunt generally includes the full body and perhaps the surround (thus the face is a smaller part of the whole image), and there may be motion blur as well.
On the other hand, neural rendering approaches currently struggle with producing ``hero'' shots where the face occupies most of a 2K or 4K image,
and they are also not suited for scenarios where artist editing is sometimes required.
\cite{Moser21,Serra22} avoid the resolution issue by using a deepfake approach only to bridge the domain gap between CG and real input images, allowing a regression from input images to PCA model coefficients to be trained with synthetic data.
%The markerless facial animation solving shows the production quality by aligning the training and unseen input footage \cite{Moser21,Serra22}.}

In summary, there is a large body of research on facial performance capture, including both fully automatic neural approaches, and approaches  that target editable high quality performance capture using classic methods. However, there is relatively little existing research on deep learning methods that target artist-in-the-loop,  editable,  and high-quality performance capture suitable for high-resolution (2K or 4K) ``hero'' facial shots \cite{Geminiman}.

\vspace{1em}
\section{
Hybrid Method for Facial Animation Capture
}
\label{sec:method}

We focus on solving performance-driven facial animation problems with blendshape models, i.e., inferring the blendshape weights $\mathbf{\hat{w}}$ such that the corresponding blendshape expression matches the captured actor's performance.
Specifically, we denote a face model by
%a column vector
$g: \mathcal{W} \rightarrow \mathcal{X}$, where $\mathcal{W} \subseteq \mathbb{R}^{D}$ is the blendshape weight space and $\mathcal{X}\subseteq \mathbb{R}^{3n}$ is the face expression space, each element of which is the $n$ vertices with the coordinates vectorized as $[x_1, y_1, z_1, \ldots, x_n, y_n, z_n]$.
Correspondingly, each blendshape is given by a vector $\mathbf{b}_k \in \mathbb{R}^{3n}$.
In the linear case a blendshape model in the "delta" formulation is: %defined as:
\begin{eqnarray}
\label{eq:bm}
g(\mathbf{w}) = \mathbf{b}_0 + \sum_{k=1}^D w_k (\mathbf{b}_k - \mathbf{b}_0)
         = \mathbf{b}_0 + \mathbf{B} \mathbf{w},
\end{eqnarray}
where $w_k$ are the blendshape weights (typically $0 \leq w_k \leq 1$), $\mathbf{b}_0$ corresponds to the neutral shape, and $\mathbf{B} \in \mathbb{R}^{3n \times D}$ contains the "delta" blendshape targets $\mathbf{b}_k - \mathbf{b}_0$.

Let us define a distance measure $\mathcal{M}: \mathcal{X} \times \mathcal{X} \rightarrow \mathbb{R}$.
Given a captured actor expression at a particular frame $\mathbf{x} \in \mathcal{X}$ ,
we seek $\hat{\mathbf{w}}$ such that $\mathcal{M}(g(\hat{\mathbf{w}}), \mathbf{x})$ is small.
% jp I worry someone could object to the "w.l.o.g", since some people feel that the entire skin must be captured i.e. sparse markers are not enough
In our case, $g(\mathbf{w})$ and $\mathbf{x}$ represent sparse markers attached on either the face puppet or the real actor's face.
In general, we can frame the problem either as a face matching problem or a regression problem.

\vspace{0.1em}  % too much space before start of sec 4
\paragraph{Matching approach.}
A prevalent approach to solving the blendshape weights is to frame it as a \emph{face matching} optimization problem \cite{Choe01}.
Choosing Euclidean distance as $\mathcal{M}(\cdot,\cdot)$ and setting $g(\mathbf{w})$ as in (\ref{eq:bm}), we can express the problem as the following optimization:
\begin{align}
\label{eq:linsolve}
\hat{\mathbf{w}} := \arg \min_{\mathbf{w}} \left\| \mathbf{b}_0 + \mathbf{B}\mathbf{w} - \mathbf{x} \right\|_{2}^{2} \quad \text{s.t.} \quad 0 \preceq \mathbf{w} \preceq 1,
\end{align}
which is a constrained quadratic programming (QP) problem.
The solution can be obtained by applying a standard QP solver run on each frame \cite{boyd2004convex,facets}.

\vspace{0.1em}  % too much space before start of sec 4
\paragraph{Learning-based approach.}
Instead of solving the face matching problem by optimization, one can also apply a regression approach to inferring the blendshape weights $\hat{\mathbf{w}}$ from the captured expression $\mathbf{x}$ through a multivariate regression function $f_\theta: \mathcal{X} \rightarrow \mathcal{W}$.

The regression function is often decomposed as $f=h \circ \phi$, where we call $\phi: \mathcal{X} \rightarrow \mathcal{H}$  the \emph{feature extractor} and $h : \mathcal{H} \rightarrow \mathcal{W}$  the \emph{regressor}.
A machine learning technique can identify %be applied to find
the function $f_{\hat{\theta}}$ given a set of labeled examples $\mathcal{D} = \{ (\mathbf{x}^{(i)}_v, \mathbf{w}^{(i)}_v) \}_{i=1}^N$ \cite{Vapnik1998}.
% such that $\mathcal{M} \left( g (f_{\hat{\theta}}(\mathbf{x}) ), \mathbf{x} \right)$ is small.
%Specifically, we consider a supervised learning setting, where $f_\theta$ is trained from a set of labeled examples $\mathcal{D} = \{ (\mathbf{x}^{(i)}_v, \mathbf{w}^{(i)}_v) \}_{i=1}^N$ drawn from a joint probability distribution $P(X, W)$.
%Given a loss function $\ell: \mathcal{W} \times \mathcal{W} \rightarrow [0, \infty)$ measuring the prediction error at one example, the supervised learning is often cast as minimizing the population risk $\mathbb{E}_{x,w \sim P}[\ell(f_{\theta}(x), w)]$. Since we only have access to the data distribution via the dataset $\mathcal{D}$, finding the optimum regression function is equivalent to minimizing the following empirical risk \cite{Vapnik1998}:
%\begin{align}
%    \hat{\theta} := \arg \min_{\theta} \frac{1}{N} \sum_{i=1}^N \ell \left(f_{\theta}(\mathbf{x}^{(i)}_v), \mathbf{w}^{(i)}_v \right).
% \end{align}
This trained regressor is expected to predict valid blendshape weights given the \emph{previously unseen} captured expression $\mathbf{x}_u$, i.e., $\hat{\mathbf{w}_u} = f_{\hat{\theta}}(\mathbf{x}_u)$, such that the 3d model expression $g(\hat{\mathbf{w}_u})$ is correctly generated.

\vspace{0.1em}  % too much space before start of sec 4
\paragraph{Nonlinear rig.}
For simplicity, the preceding description presents the matching and learning-based approaches in terms of a linear blendshape model.
Our blendshape puppets incorporate natural non-linear expressions, for example using in-between and corrective shapes \cite{facuda,osipa10}. In addition, the overall puppet consists of a large deformation chain with numerous tweaks, deformers and skin clusters applied on top of the blendshape system.

 A full description of the puppet is outside the scope of this paper. The details of the puppet are not needed to understand the application of the learning and matching approaches, however:  the learning-based approach captures the puppets's behavior (\emph{including} nonlinearities) since the puppet itself is used to produce training data (Section~\ref{subsec:syndata}).
For the matching approach, rather than seek a global optimum, our optimization fine-tunes a local linear subset of parameters that are selected by the artist (see \emph{Hybrid approach}, next, and Section~\ref{subsec:finetune}).
%  Note that \eqref{eq:bm}, \eqref{eq:linsolve} describe the linear case for simplicity. In the case of a rig with nonlinear in-between shapes \cite{osipa10,lewis-blendshape-2014} a nonlinear minimization is used \cite{Liu1989}.

\vspace{0.1em}  % too much space before start of sec 4

\paragraph{Hybrid approach.}
In FDLS, we apply the learning-based approach as the \emph{main animation solver}, specifically using deep learning regression \cite{he-resnet-2016}, followed by the matching approach to fine-tune some facial parameters.
This hybrid approach has two main advantages.
First, the main animation solving is reduced to the \emph{forward pass} of the network $f_{\theta}$, which is time-efficient.
Second, the forward pass provides an initialization for the nonlinear matching optimization that is in the neighborhood of the correct local minimum and is close to the desired solution, resulting in temporally consistent and efficient fine tuning to generate high-fidelity final animations.
In the actual implementation, FDLS comprises several components to enable the \emph{human-in-the-loop} solving process. These are fully elaborated in the next section.

%\vspace{-0.1em}
%\section{The Complete Pipeline of FDLS} \label{sec:pipeline}
\section{The Complete FDLS Pipeline} \label{sec:pipeline}

%        _            _ _
%  _ __ (_)_ __   ___| (_)_ __   ___
% | '_ \| | '_ \ / _ \ | | '_ \ / _ \
% | |_) | | |_) |  __/ | | | | |  __/
% | .__/|_| .__/ \___|_|_|_| |_|\___|
% |_|     |_|
\begin{figure*}
    \centering
    \includegraphics[width=\textwidth]{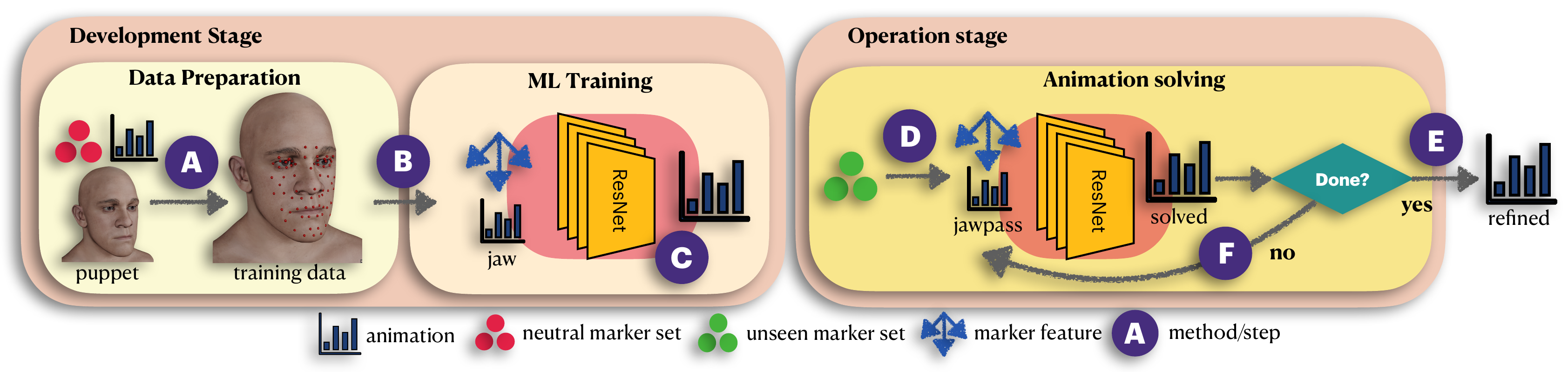}
%   \caption{The Facial Deep Learning Solver (FDLS) comprises three main stages: data preparation, training and solving. FDLS pipeline is shown on the left, where the letters denote:
    \caption{The Facial Deep Learning Solver (FDLS) comprises three main stages: data preparation, training and solving, where the letters denote: (A) salient sample selection (\Secref{subsec:salient}); (B) feature engineering (\Secref{subsec:feature}); (C) jaw kinematics conditional training (\Secref{subsec:jawcond}); (D) shape alignment (\Secref{subsec:alignment}); (E) finetuning (\Secref{subsec:finetune}); (F) anchor poses (\Secref{subsec:alignment}).}
    \label{img:pipeline}
\end{figure*}

%In general,
As shown in \Figref{img:pipeline}, the end-to-end FDLS pipeline consists of two main stages: i) the development stage, and ii) the operation stage.
This partitioning forms a \emph{human-in-the-loop} MLOps lifecycle in Weta Digital's production that ensures reproducibility, testability, and maintainability of our deep learning system.
In the development stage, data preparation steps are performed, including synthetic data generation (\Secref{subsec:syndata}), salient sample selection (\Secref{subsec:salient}), feature engineering (\Secref{subsec:feature}), and training the deep learning models (\Secref{subsec:dl}).
The animation solving happens in the operation stage, which makes use of the trained deep learning models as the main solvers complemented by a few other techniques: shape alignment with anchor poses (\Secref{subsec:dl}) and fine-tuning optimization (\Secref{subsec:finetune}).

\subsection{Synthetic Data Generation and Augmentation} \label{subsec:syndata}

% maybe not necessary to say "sparse" markers, the markers are always(?) sparse
The training data for FDLS are in the form of 3d virtual markers created from the actor's facial landmarks of his/her neutral expression using photogrammetry -- we call these "neutral markers" throughout the paper for brevity.
We then attach the neutral markers to the corresponding actor's blendshape puppet and synthetically generate the training markers driven by the puppet motion $\mathbf{x}^{(i)}_v = g(\mathbf{w}^{(i)}_v)$ forming a set of training tuples $\mathcal{D} = \{ (\mathbf{x}^{(i)}_v, \mathbf{w}^{(i)}_v) \}_{i=1}^N$.
Note that the training blendshape weights $\mathbf{w}^{(i)}$ are
designed and selected %flexibly hand-crafted
by animation experts,
typically containing both FACS and some actor-specific expressions.
Since the training markers are driven by the puppet, the training data thus implicitly includes nonlinear effects from the puppet without requiring a differentiable chain of deformations.

Additionally, we also utilize a data augmentation or oversampling technique to further increase the variation of the synthetic markers $\mathbf{x}_v$ and therefore avoid overfitting of the trained deep learning models.
This is achieved by randomly injecting Gaussian noise in the original markers $\mathbf{x}_e = \mathbf{x}_v + \mathcal{N}(0, \Sigma)$.
We empirically determine the covariance $\Sigma$ by analyzing the jitter of the tracked markers,
for instance, a marker at the nose bridge has less variance than those on the chin region.
This can produce a significant quality gain in the solved animations.

\subsection{Salient Training Sample Selection} \label{subsec:salient}

As is mentioned in \Secref{subsec:syndata}, the training data for FDLS can be flexibly generated with various combinations of simple
% We can train using a simple set of examples, e.g.,
``one-hot'' single action unit FACS expressions (producing a minimum viable solution) and more complex and realistic data from various sources such as dynamic scans of the actor,
previously animated or solved performances, etc.

While we generally expect that more realistic and complex training data should result in better performance, counterintuitively we observed that this is not always the case. We found that the root cause is data imbalance -- for example, some solved performances may consist mostly of a neutral expression.
This issue can be avoided if the additional training examples are carefully selected by the animators, but manually selecting from among many training expressions is laborious.

To simultaneously encourage data diversity and salience,
we introduce a salient sample selection technique to automatically remove redundant examples from the training set.
This eliminates the need for tedious manual sample selection. %from among a many training expressions.
We reduce the number of training examples from $N$ to $M < N$ by sequentially applying the following sample selection rule
\begin{equation}
\label{eq:salient}
    s_i = \frac{1}{N}\left( \sum_{j=1}^{N} k( \xx^{(i)}, \xx^{(j)} ) \right) < \sigma,
\end{equation}
where $s_i \in \{0,1\}$ is a binary variable indicating whether the shape $i$ is selected, $\sigma$ is a tunable threshold, and $k: \mathcal{X} \times \mathcal{X} \rightarrow [0,1]$ is the similarity measure.
In the extreme case, this salient sample selection picks only a single datapoint from the dataset $\mathcal{D}$ if it contains mostly similar expressions according to Eq.~\eqref{eq:salient}, e.g., neutral non-dialogue expressions.
We choose an exponential Radial Basis Function (RBF) \cite{Bro88} representing $k(\cdot, \cdot)$ which provides a good set of salient shapes due to its effectiveness in capturing similarities in a non-linear feature space.

%\vspace{-1em}
\subsection{Feature Engineering} \label{subsec:feature}

%Using solely the sparse marker coordinates
Solving animation using solely sparse marker coordinates $\xx = [\mm_1, \ldots, \mm_n]$, where $\mm_i \in \mathbb{R}^3$ to represent faces %as the basis to solve animation
has its own limitations (e.g., \cite{seolactorspecific}).
Sparse markers are less informative than the full face representation (e.g., facial depth from LiDAR), and it is not possible to fully characterize certain facial expressions using only marker coordinates.

% \textbf{In particular, when the jaw is only slightly open it is difficult to determine if the teeth are visible.}
To partially address this and simultaneously help with the limited amount of training data available in this domain,
we developed several \emph{graph features} to increase the information richness of the input representations, thereby inducing robust and highly accurate deep learning solves.

In the initial enthusiasm following the success of \cite{alexnet}, it was argued that features emerging from neural net training should outperform manually designed features aka "feature engineering".
While this is arguably true especially in computer vision and natural language processing domains, engineered features continue to be used in state-of-the-art deep learning -- the positional encoding in transformers and Fourier/positional encoding that distinguishes NeRF \cite{mildenhall2020nerf} are prominent examples.
In practice, the need for engineered features can be justified when it is impractical to search across a sufficient variety of architectures to find those that result in ideal features, or to impose an inductive bias in cases where "correct" features might not be discovered due to limited data -- as is often the case in visual effects!

\begin{figure}[hbt]
    \centering
    \includegraphics[width=0.8\linewidth]{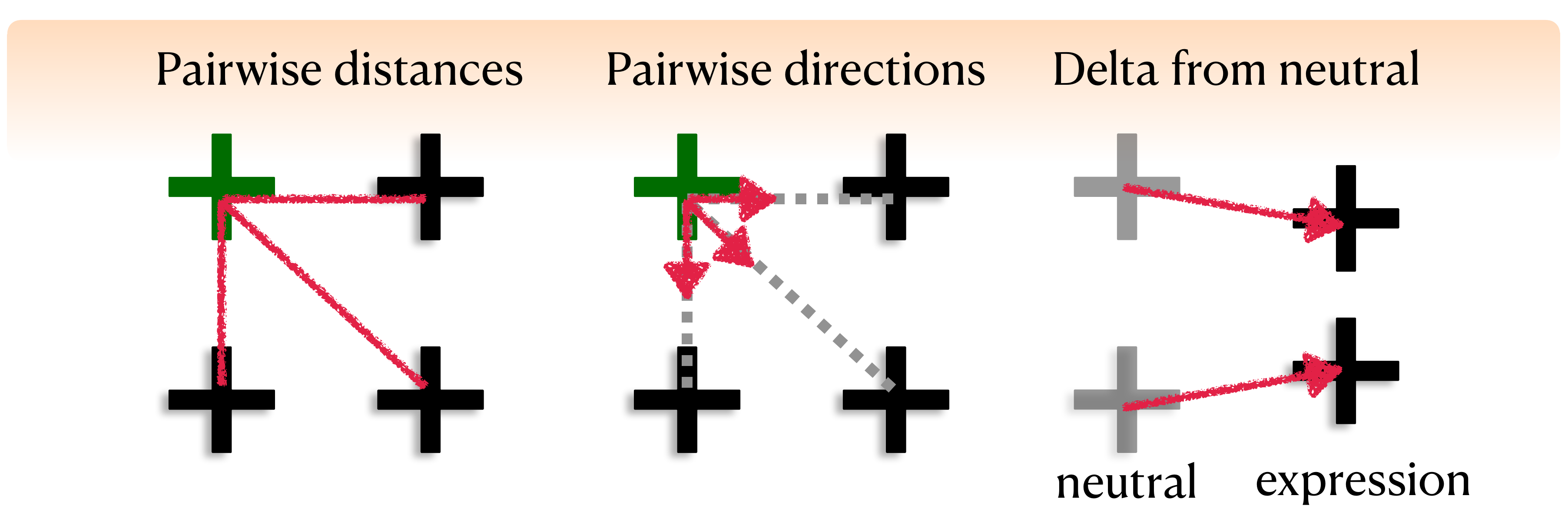}
    \caption{FDLS graph features. From left to right: pairwise distance, pairwise directions, delta from neural to the target expression.}
    \label{img:feature}
 \vspace{-1em}
\end{figure}

FDLS benefits from carefully engineered features extracted from the marker coordinates.
\Figref{img:feature} illustrates the variants of the feature extractor $\phi: \mathcal{X} \rightarrow \mathcal{H}$ that we design as part of the learning-based approach (recall \Secref{sec:method}). These are described next. %described as follows.

% % % pairwise distance feature
\paragraph{Pairwise distance.}
The pairwise distance features $\boldsymbol{\phi}^{\text{dist}} \in \mathbb{R}^{n^2}$ are a feature vector, whose values represent
the Euclidean distances over all possible combinations of two marker coordinates in the input representation $\{\mm_i\}_{i=1}^n$.
The $k$-th element of the pairwise distance feature $\phi^{\text{dist}}_k \in \mathbb{R}$ is given by $\phi^{\text{dist}}_{k}  = \norm{ \mm_i - \mm_j }_2.$
% \begin{equation}
%     \boldsymbol{\phi}^{\text{dist}}_{k}  = \norm{ \mm_i - \mm_j }_2.
% \end{equation}

% % % pairwise directional feature
\paragraph{Pairwise direction.}
The pairwise direction features $\boldsymbol{\phi}^{\text{dir}} \in \mathbb{R}^{3n^2}$ are a collection of the directional vectors from one coordinate to another.
This gives complementary information to eliminate the inherent ambiguity in $\boldsymbol{\phi}^{\text{dist}}$.
For instance, the Euclidean distance of two particular coordinates between different expressions, e.g. neutral and smiling, could be indistinguishable due to symmetry.

Analogous to the pairwise distance computation, the $k$-th coordinate $\boldsymbol{\phi}^{\text{dir}}_k \in \mathbb{R}^3$ is computed as $\boldsymbol{\phi}^{\text{dir}}_k = (\phi^{\text{dist}}_k)^{-1}(\mm_i - \mm_j).$
% \begin{equation}
%     \boldsymbol{\phi}^{\text{dir}}_k = \frac{\mm_i - \mm_j}{\phi^{\text{dist}}_k}.
% \end{equation}
Note that the pairwise distance and direction features are both invariant to small slippage of the head camera.

\paragraph{Delta pose.}
Lastly, we define the delta pose features $\boldsymbol{\phi}^{\text{delta}} \in \mathbb{R}^{3n}$, which are the coordinate differences between a particular face expression and its neutral pose.
These features are invariant to the global displacement of the marker coordinates and have become a common representation used in traditional blendshape systems in the graphics community \cite{lewis-blendshape-2014}. % removed jp2010, 2014 is enough
Given the specific expression $\xx$ and the corresponding neutral pose $\xx_{0}$, the delta pose feature vector $\boldsymbol{\phi}^{\text{delta}} \in \mathbb{R}^{3n}$ is calculated as $\boldsymbol{\phi}^{\text{delta}} = \xx - \xx_{0}$.
% \begin{equation}
%     \boldsymbol{\phi}^{\text{delta}} = \xx - \xx_{0}.
% \end{equation}

The complete set of input features for FDLS is the concatenation of these three features, i.e., $\boldsymbol{\phi} = [\boldsymbol{\phi}^{\text{dist}}, \boldsymbol{\phi}^{\text{dir}}, \boldsymbol{\phi}^{\text{delta}}] \in \mathcal{H}$.
Intuitively, these features are a graph-like representation of facial expressions that contains not only the positional information but also pairwise relationships over the marker coordinates. These features can be directly applied in our region-based training by extracting the features only in a specific face region, as will be shown later.
Graph features have been used in concurrent work such as \cite{qi-pointnetplus-17}.
% DGCNN could be another citation
Our use is somewhat distinguished in that we intentionally do not introduce convolution. The translational equivariance of convolution is appropriate for recognition of general point clouds (for example an edge may appear at any location) however the features of a canonically positioned face are in relatively fixed locations and we seek only to capture changes due to expression.

% \subsection{Jaw Kinematics Conditional Training} % \label{subsec:jawcond}

% omitted "DL" in the subsection title - think DL is not defined
% and it is not needed because the first sentence says "deep learning"
\subsection{Region-based Training and Solving}
\label{subsec:dl}

The deep learning regression $f_{\theta}$ defined in \Secref{sec:method} is the core of FDLS. %which is developed through the learning-based approach as described in \Secref
In its implementation, we define seven deep residual networks, each responsible for a particular face region containing a subset of the sparse markers.
We denote the face regions by $R \in \{\textit{upper-face}, \textit{lower-face}, \textit{jaw}, \textit{lips}, \textit{cheek}, \textit{eye-lids}, \textit{eyeballs} \}$.
The marker subset is selected according to the semantic meaning of the face region, but a marker on the nose bridge is always included regardless of the region to insure sufficiently diverse and spatially supported features. %to expose more variations in the feature space.
Each face region is also linked only to a subset of blendshape weights or channels that are relevant to that region.
In other words, we train $\{ f_{\theta_r}: \mathcal{X}_r \rightarrow \mathcal{W}_r; \forall r \in R \}$.
%in a supervised setting using Adam gradient-based optimization \cite{kingma:adam}.  % suggest move this to the supplementary together with the description of number/width of layers

Solving with the trained deep learning models is straightforward, i.e., we simply run the forward pass $\hat{\ww}_r = f_{\hat{\theta}_r}(\xx_r)$ for each face region on a per frame basis.
The full solved weights are the concatenation of $\{ \hat{\ww}_r \}_{r \in R}$, which we refer to as the \emph{raw animation} -- later this will be used as the basis for editing or fine-tuning. As described next, a substantial improvement in the quality of the raw animation is obtained by extending the forward pass with two additional techniques.
% Keep for history

\begin{figure}[H]
    \centering
    \includegraphics[width=\linewidth]{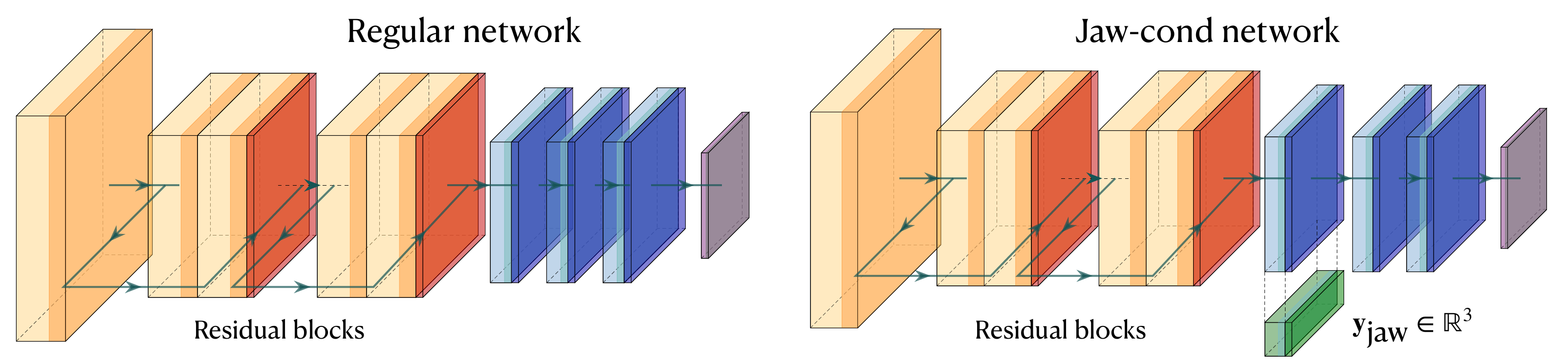}
    \caption{
    The network $f_{\hat{\theta}_r}(\xx_r)$ for a regular face region (left) and a jaw-conditioned network  $f_{\hat{\theta}_{l}}(\xx_{l}; \hat{\ww}_{\text{jaw}})$ (right).
    % "used in FDLS training" -> i think used in inference also(?)
    }
    \label{img:network}
\end{figure}

\subsubsection{Jaw Kinematics Conditional Training.}
\label{subsec:jawcond}
Jaw kinematics plays an important role in affecting the shape and expressions of the face, especially the lower half~\cite{Yang19}.
For instance, one can have different smiling expressions under different jaw conditions (open, close, sideway movement, etc).
In other words, the effect of face muscle activations controlling the lips and cheeks strongly depend on the jaw position.

% I assume you mean a standard at weta. if it is a standard in the industry
% probably it should have a citation
For this reason, in our pipeline the jaw motions are established \emph{before} solving other regions. FDLS supports the artist in this process by providing a regression specifically for the jaw position.
Specifically, we denote the jaw activation weights by $\ww_{\text{jaw}} \in \mathbb{R}^z$ where in our production face models, normally $z=3$, namely, \texttt{jawOpen}, \texttt{jawThrust}, and \texttt{jawSideways}.
The jaw motions $\hat{\ww}_{\text{jaw}}$ are obtained from computing $f_{\hat{\theta}_{\text{jaw}}}(\xx_{\text{jaw}})$
and are used in the following regions $l \in \{\emph{lower-face}, \emph{lips}, \emph{cheek} \}$. We solve the animation by evaluating $f_{\hat{\theta}_{l}}(\xx_{l}; \hat{\ww}_{\text{jaw}})$, which is trained by the risk minimization:
\begin{equation}
    \displaystyle
    \hat{\theta}_{l} := \arg \min_{\theta_{l}} \frac{1}{N}
    \sum_{i=1}^N \ell \left(f_{\theta_{l}} (\xx^{(i)}_{l}; \ww^{(i)}_{\text{jaw}}), \ww^{(i)}_{l} \right),
\end{equation}
% \TODO{Mistake, all neural network input should be features $\phi$, rather than marker $\xx$. We probably need to fix the training tuple mentioned in Sec4.1 as well}

where $\ell: \mathcal{W} x \mathcal{W} \rightarrow \mathbb{R}$ is the standard regression loss function.
Note the use of $\ww_{\text{jaw}}$ (the groundtruth jaw motion) in the training, versus $\hat{\ww}_{\text{jaw}}$.
% Note that $\ww_{\text{jaw}}$ is the groundtruth jaw motion for training,  different from $\hat{\ww}_{\text{jaw}}$.

As we can see the solved jaw weight vector $\hat{\ww}_{\text{jaw}}$ is part of the inputs to the deep learning model $f_{\theta_{l}}$.
However, due to its low dimensionality, we embed it as additional nodes in a hidden layer after the series of deep residual blocks rather than as part of the input layer, as shown in \Figref{img:network}.

\subsubsection{Shape Alignment with Anchor Poses.}
\label{subsec:alignment}
In real productions lasting several days or more, the positions of markers on a single actor's face may shift over time. There are several reasons for this. First, although the landmarks are painted using a perforated mask and their relative positions do not change, the mask may be placed in slightly different positions on different days. Second, the accuracy of the tracked and 3d-reconstructed markers can be inconsistent across different shots, depending on their complexity. Lastly, the actor may slightly gain or lose weight. These changes can introduce a small shape mismatch between the puppet-snapped markers and the performance-captured markers. %(see \Figref{img:intpose},

\begin{figure}[H]
    \centering
    \includegraphics[width=0.8\linewidth]{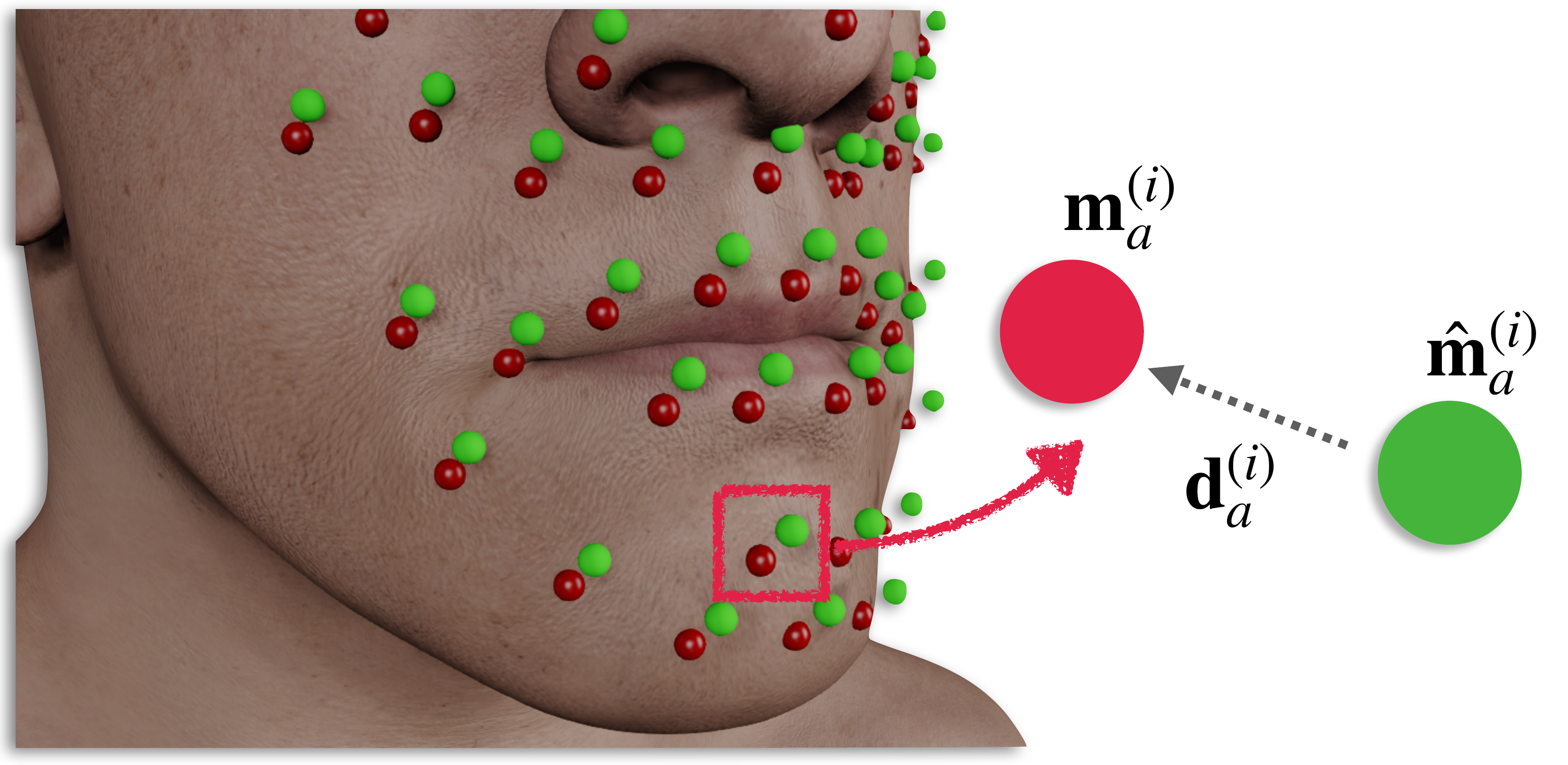}
    \caption{Shape alignment using an anchor pose. The green and red circles represent the puppet-snapped markers and the performance-captured markers, respectively.
    }
    \label{img:intpose}
 \vspace{-1em}
\end{figure}

To mitigate this problem, we introduce a simple shape alignment method that can be performed directly by an animator before (re)running the deep learning solver $f_{\hat{\theta}}(\cdot)$.
The basic idea is to first calculate the offsets between the synthetic markers and the shot-captured markers producing an \emph{anchor pose}, i.e., a hand-picked expression at a particular frame in a shot, and then % linearly  only the first one is linear
propagate the offsets to displace the other face markers within the entire frame range.
Specifically, suppose that we have a performance shot with $T$ frames and an anchor pose at frame $a$ is chosen.
The artist then sets the anchor weights $\ww_a$ such that $g(\ww_a)$ portrays the desirable anchor pose.
The shape alignment is executed as follows:
\begin{equation}
\begin{split}
    & \dd_a = \xx_a - g(\ww_{a}), \\
    & \tilde{\xx}_f = \xx_f - \qq_{fa} \odot \dd_a,\ f \in \{1,...,T\},
\label{eq:anchor}
\end{split}
\end{equation}
where $\dd_a$ are the marker offsets and $\qq_{fa} \propto \exp{(-\|\xx_a - \xx_f\|^2_2)}$ denotes the weighting factors of the alignment by calculating the pose similarity.
% jp in this equation I am guessing that the argument to exp is scaled somehow, like exp(-(1/c)||xx_a - xx_f||^2)? To account for that, instead of "=" it could be "\propto" to indicate there is a not-interesting scale
After this we can rerun the deep learning solver $\tilde{w} = f_{\hat{\theta}}(\tilde{\xx}_f), \forall f= \{1, \ldots, T \}$ \emph{without retraining} to produce a more accurate animation.
Note that this shape alignment process can be done iteratively by choosing a few other anchor poses as needed -- for the first anchor pose, all elements of $\qq$ are set to one.

\subsection{Fine-tuning Optimization} \label{subsec:finetune}

Editing can optionally be performed once the raw animation has been solved by the deep learning model $\hat{\ww}_u = f_{\hat{\theta}}(\xx_u)$.
In cases with unusual expressions, we sometime see a small discrepancy between the solved expression of the actor puppet $g(\hat{\ww}_u)$ and the actual captured expression $\xx_u$, mainly because the shape $\xx_u$ has not been well represented in the training dataset for constructing $f_{\hat{\theta}}$. The artists can simply adjust the solved weights (most of the time only a few channels are needed) to get more accurate expressions.

To make the aforementioned process more scalable, we provide a \emph{fine-tuning optimization} as a post-processing step in FDLS.
It essentially further minimizes the error $\| g(\hat{\ww}_u) - \xx_u \|_2$.
% jp we are not clear about this point.  eq:bm is linear, however I believe the puppet is not, due to the combination shapes, which are not represented in eq:bm. We can revisit the truth later, but for the immediate deadline it's probably best to leave out the linear statement here
We run the minimization by solving the face matching problem in Eq.~\eqref{eq:linsolve}, in which the deep learning solved weights are set as the starting point for a minimization using  the L-BFGS algorithm as the optimizer ~\cite{Liu1989}. We found this provides better results than starting the minimization from zero or a random point.
% jp the part "are conditioned on the other solved weights" -- I know what you mean, but I think strictly this isn't true, the minimization just minimizes the error over the subset of weights and does not know the value of the other weights
Note that we also form a restricted blendshape basis (matrix $\mathbf{B}$ in Eq.\eqref{eq:linsolve}) that only includes artist-selected degrees of freedom to be finetuned while the remaining weights remain frozen.
% \Figref{img:jawcondface} shows examples of the specialized basis shapes that compose the matrix $\mathbf{B}$.

% \begin{figure}[H]
%     \centering
%     \includegraphics[width=\linewidth]{assets/fdls-finetune-basis.pdf}
%     \caption{Examples of a specialized basis for the finetuning optimization. From left to right: raw expression from deep learning model, followed by bases corresponding to the selected shapes to be finetuned:  lipFunneler, lipCornerPuller, chinRaiser, and upperLipRaiser, respectively.
%     }
%     \label{img:jawcondface}
% \end{figure}

With this implementation, we provide the artists with flexible options
%several degrees of freedom
such as choosing only a specific frame range, partially selecting the face markers as objects of comparison, and selecting only a few blendshape weights to be finetuned.
This makes the fine-tuning optimization an interactive tool as part of an overall editing flow to efficiently finalize high quality animations.

\section{Experiments}

In this section we review the end-to-end FDLS pipeline and present the main results as well as ablation tests.
While the use case shown here is for a specific actor and a single shot performance, the workflow is generic for any actor and performance capture.

% single image
\begin{figure}
    \centering
    \includegraphics[width=.5\textwidth]{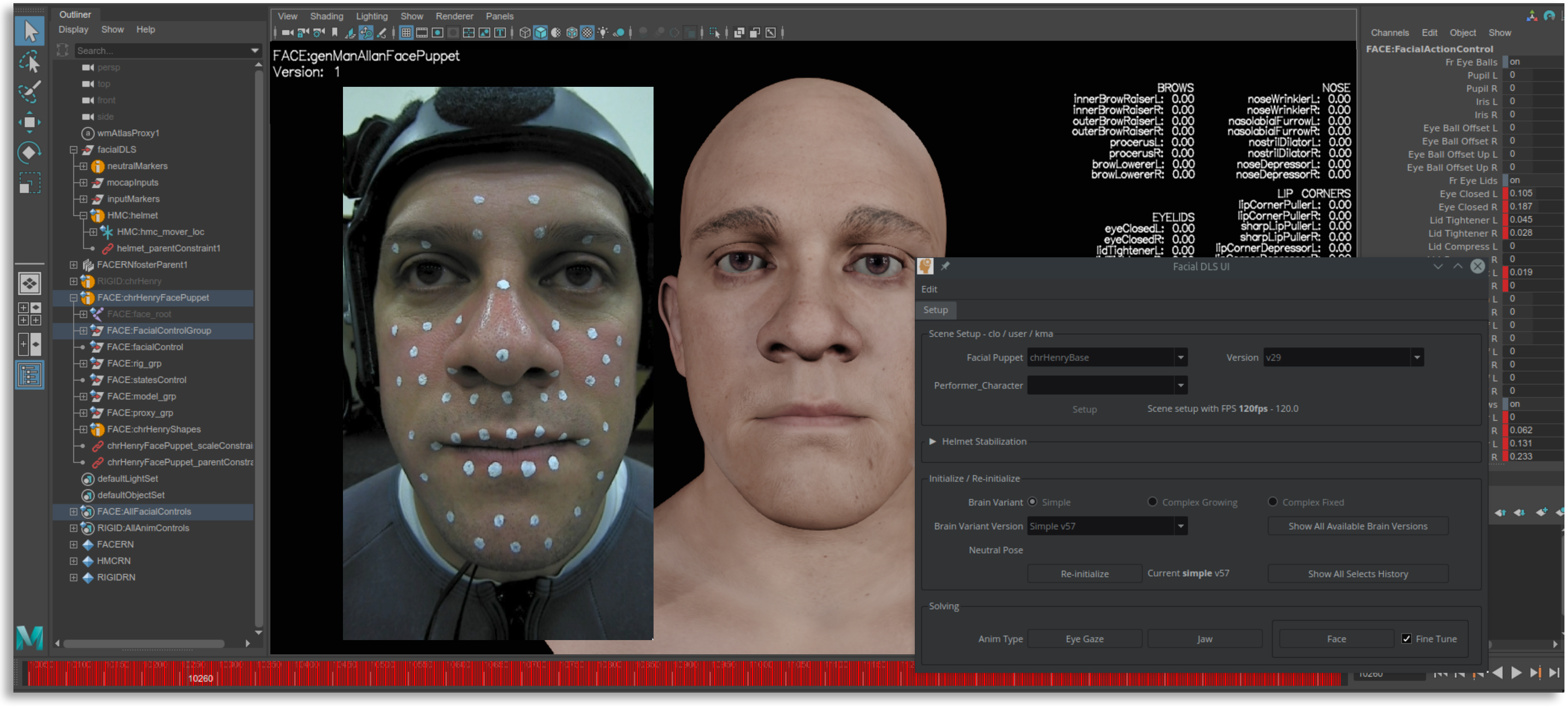}
    \caption{FDLS workspace and GUI. The system is implemented using the commercial Maya API \cite{maya}.}
    \label{img:solve:workspace}
\vspace{-0.3cm}
\end{figure}

% % % two image side by side
% \begin{figure*}
%     \centering
%     \includegraphics[width=\linewidth]{assets/fdls-anchor.pdf}
%     \caption{The different performance of the same clip, where some in-between frames are skipped. Top: The final performance edited by an artist. Middle: The raw solve of FDLS without anchor pose. Bottom: The finetune with one anchor pose highlighted with red rectangle. \TODO{To replace fourth column; To rearrange the row as rawsolve/finetuned/animator result}}
%     \label{img:exp:anchor}
% \end{figure*}

% % two image side by side
\begin{figure*}
    \centering
    \includegraphics[width=\linewidth]{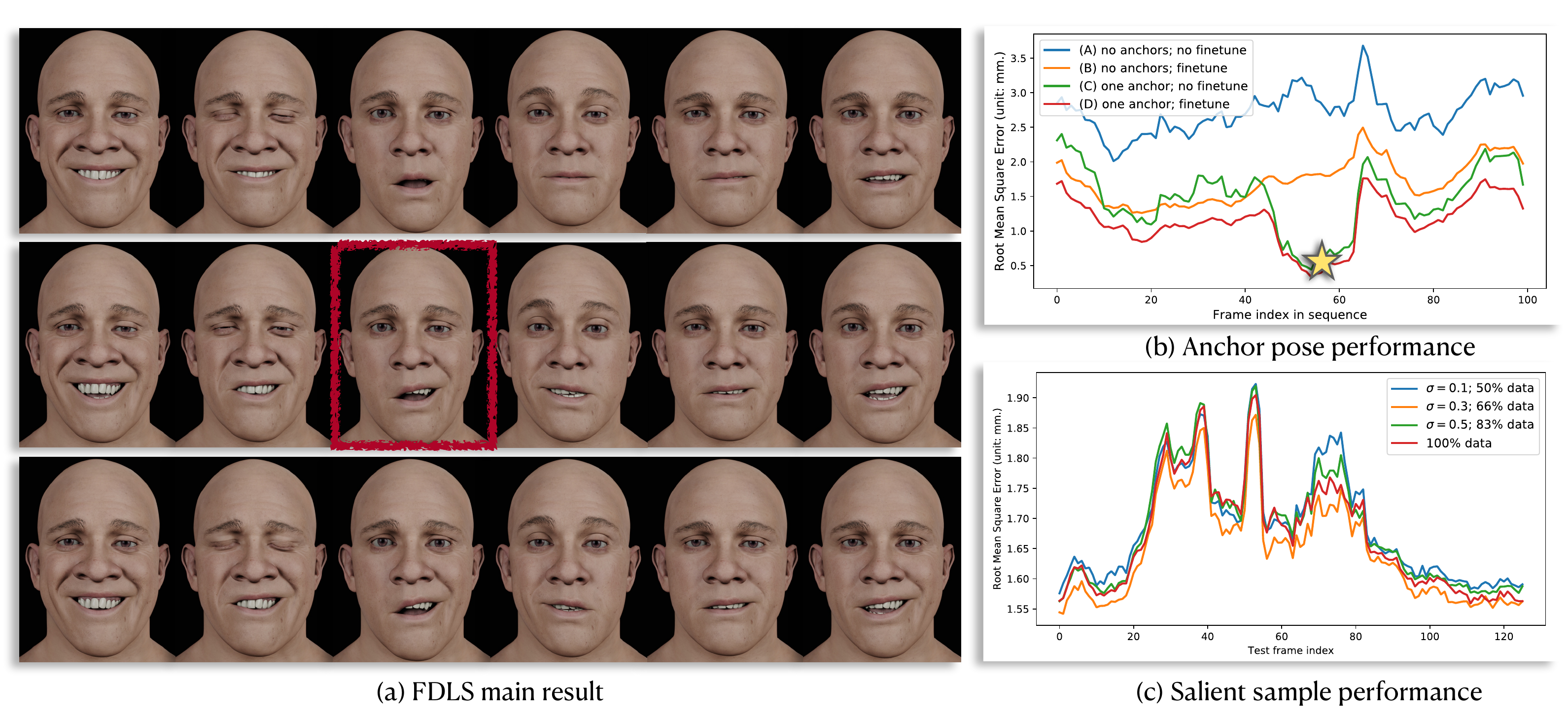}
    \caption{A complex FDLS solve example. (a) %from top to bottom: says top/middle/bottom so from top to bottom not needed
    raw solve with shape alignment from a single neutral-ish frame (top), finetune with the additional anchor pose outlined in red (middle), and the final manual edits from artist on top of the FDLS solve (bottom). (b) and (c) are the RMSE performance of ablation experiments at the frame marked with star with/without an anchor pose and finetuning, and on the salient data selection, respectively.
    Note that the RMSE generally includes an irreducible offset due to slight mismatch between the actor and puppet virtual  marker positions.
    }
    \label{img:exp:anchor}
\end{figure*}

%  __  __       _
% |  \/  | __ _(_)_ __
% | |\/| |/ _` | | '_ \
% | |  | | (_| | | | | |
% |_|  |_|\__,_|_|_| |_|

% \subsection{Main results} \label{subsec:result}
% jp could remove this sentence if needed to save space:
Next we describe the implementation details of the FDLS training and solving stages as depicted in \Figref{img:pipeline}.

%  _____          _
% |_   _| __ __ _(_)_ __
%   | || '__/ _` | | '_ \
%   | || | | (_| | | | | |
%   |_||_|  \__,_|_|_| |_|

\subsection{Training}

As described in \Secref{subsec:syndata}, we synthetically manufacture the training data in the form of a series of sparse markers by assigning animation weights to the blendshape channels. In the FDLS training, two types of training animations are utilized:  ``one-hot'' FACS expressions and the range of motion (ROM) of an actor.

The one-hot FACS animation has individual FACS expressions such as jaw drop (AU26) and lip corner puller (AU12) at successive frames, thus representing the full extent of the blendshape basis.

By using only the one-hot FACS training animation, we can already construct a \emph{minimum viable solver} before more advanced training animation is available.

The ROM captures the motion of the specific actor. This is more realistic than the one-hot animation, as some people have difficulty activating individual muscles in isolation. One can think of the FACS-based animation as  determining the range and extremes of individual muscles, while the ROM provides information about their distribution, for example, what muscle combinations actually occur and do not occur for the particular actor. In addition, the ROM allows an expert artist to specify exactly which blendshape weights contribute to
an expression, thus guiding the solver to disambiguate cases where nearly (but not exactly) identical marker positions can be produced with different weight configurations.

%  ____        _
% / ___|  ___ | |_   _____
% \___ \ / _ \| \ \ / / _ \
%  ___) | (_) | |\ V /  __/
% |____/ \___/|_| \_/ \___|
% \subsubsection{Solving}
\subsection{Solving}

% (save space) this message is in the conclusion, not necessary to also say it here I think
\Figref{img:solve:workspace} shows the FDLS workspace and GUI. Note that we do not require the artist to specify any parameters to perform solving, thus the tool is convenient to use without significant artist training. Our users have reported that FDLS system takes only about one week to train a new animator, which boosts the production speed.

After the workspace has been setup (\Figref{img:solve:workspace}), FDLS allows the artist to pick a neutral-ish frame
% jp because the description was changed and doesnt say"pose initialization" now
as an anchor frame (\Secref{subsec:alignment}), and correct the jaw animation given the raw jaw solve, then run the jaw-conditioned networks (\Secref{subsec:jawcond}). After raw solve, one or more additional anchor poses can be added to refine the projected markers. Finally, the post-finetuning optimizes the face conditioned on jaw channels to match the performance markers.

\Figref{img:exp:anchor}(a) shows a complex FDLS solve that illustrates most aspects of the system. The top row depicts the raw solved expressions induced by the region-based models $\{ f_{\theta_r}(\cdot) \}_{r \in R}$ after shape alignment using one neutral-ish frame. The middle row displays the solved expressions with shape alignment using the additional anchor pose highlighted in red followed by finetuning. This shows a substantial improvement in that the result is close to the final desired expressions (bottom).

\Figref{img:exp:anchor}(b) compares the performance of post-finetuning versus the raw solve, with or without an anchor pose at the marked frame. The minimization ensures the RMSE loss value decreases (e.g., curves A and B), however the error remains high across frames in this case. By adding one anchor pose at around frame 55 (shown in (a) with the red box), the error is globally decreased with no further effort.

\Figref{img:exp:anchor}(c) shows the results of salient frame selection by varying the threshold $\sigma$ in \eqref{eq:salient}. Specifically, it compares the test performance with $\sigma=0.1$, $\sigma=0.3$, $\sigma=0.5$, and the full dataset, corresponding to using $50\%$, $66\%$, $83\%$ and $100\%$ of the original dataset, respectively. One can see that the network trained with data selected using $\sigma=0.3$ (orange) has better performance than the others, including the network trained with the full dataset (red).

Many performances include a certain amount of repeated information, for instance, in some performances a large percentage of the frames show a mostly neutral expression. The experiment in \Figref{img:exp:anchor}(c) provides evidence of an important point, which is that the solve actually \emph{improves} when this redundant data is removed. However, a limitation of this approach is that the $\sigma$ must be empirically tuned for the particular dataset. If $\sigma$ is set to an extreme (e.g., $\sigma=0.1$) or moderate value (e.g., $\sigma=0.5$) it may give poor results.

\section{Conclusion and future work} \label{subsec:future}

This paper presents a human-in-the-loop deep learning based approach to facial animation solving. The FDLS system is a pioneering application of deep learning to facial solving in movie production.
It was initiated in 2016 and has seen ongoing development and improvement since then.
It presents a simple and light-weight training and solving pipeline to our artists. The approach is suitable for driving sophisticated and non-linear blendshape rigs with hundreds of parameters. FDLS produces production quality results while requiring limited training data and artist effort.
%Both
The design trade-offs and several specific features of the system are guided by consideration of artist practice and differ from other published work.

The graph features have proven successful at allowing accurate solves with limited training data.
% HMC = head mounted cam?  "emotion challenge" has some term. (Yes, head mounted cam)
The quality of the intermediate solution can be verified and  adjusted if needed following both the jaw estimation and prior to the post fine-tuning, and since FDLS produces interpretable parameters, the freedom to do further editing on the final solution is fully preserved.
The anchor pose mechanism allows the network to adapt to slight changes in both the marker positions and the actor's face shape without re-training.

In our experience animators can learn to use the system in approximately a week % CHECK/FIX THIS, how many hours or days?
and a large majority of shots require little or no human editing. While we could cherry pick results to provide an ``objective'' characterization of the production performance, in all honesty such a characterization is almost meaningless. This is due to wide variety of different performance and retargeting situations, the different ways that FDLS can be used, the quality of training data, choice of anchor poses, and other factors, and particularly because of the ultimate perceptual evaluation of the results. Instead, we highlight that FDLS succeeds as a system that can be employed in \emph{all} these scenarios -- from fully automatic solves to extremely challenging shots that require iterative solution and evolving guidance from a supervisor. %We explicitly list our contributions in the following:

\iffalse

\begin{itemize}
    \item The anchor/initialization pose prevents the risk to re-train the neural network whenever there's a difference between the training basis and unseen data.
    \item The feature engineering augments the dimensions of sparse marker set producing better result from wider neural network configuration.
    \item Jaw-conditioning training and post-finetuning shed light on the final quality.
\end{itemize}
\fi

One of the disadvantages of using a sparse marker set is the ambiguity of the lip positions in certain expressions. In the future we would like to enhance the solver with additional information such the lip contours or LIDAR depth.

%\section*{Acknowledgments} \label{subsec:ack}
\begin{acks}
We thank Joe Letteri, Marco Revelant, Luca Fascione, Dejan Momcilovic, Stephen Cullingford, Stuart Adcock, Allison Orr, David Luke, Millie Maier, Andrew Moffat, Kenneth Gimpelson, Nivedita Goswami, and Zhicheng Ye for supporting this project. Moreover, we appreciate the anonymous reviewers for their suggestions.
\end{acks}

%\begin{acks}
%To Robert, for the bagels and explaining CMYK and color spaces.
%\end{acks}

\bibliographystyle{ACM-Reference-Format}
\bibliography{main}

% \appendix

% Document starts
% Title portion
% \newpage
% \onecolumn
%\begin{minipage}{\textwidth}
%\newpage
%\begin{center}
%    \LARGE \textbf{Supplementary Material}
%\end{center}

% \title{FDLS: A Deep Learning Approach to Production Quality, Controllable, and Retargetable Facial Performances: Supplementary Material}

% randaugment

%\maketitle

%\end{minipage}
% \twocolumn
% \title{FDLS: A Deep Learning Approach to Production Quality, Controllable, and Retargetable Facial Performances: Supplementary Material}
% \keywords{}
% \begin{abstract}
% \end{abstract}
% \maketitle

%\onecolumn\newpage\twocolumn
\clearpage

\begin{strip}
\begin{center}
    \Huge\textbf{FDLS: Supplementary Material}
    %\Huge\textbf{FDLS: A Deep Learning Approach to Production Quality, Controllable, and Retargetable Facial Performances: Supplementary Material}
\end{center}
\end{strip}

\section{Data Acquisition}

The data acquisition step is crucial to generate the FDLS training tuples $\mathcal{D} = \{ (\mathbf{x}^{(i)}_v, \mathbf{w}^{(i)}_v) \}_{i=1}^N$. It uses the following three components: neutral markers, the facial puppet, and facial motion (aka blendshape weights).

The neutral markers are created in a motion capture session by triangulating from images of the actor's neutral expression captured through multiple cameras.
%It is generated by the triangulation process of actor's neutral expressions captured through multiple cameras.
To make the static neutral markers animatable, we project each marker on the facial puppet.

The facial motion is created by the artist and comprises FACS expressions and range of motion (ROM). The FACS motion contains about 500 frames. This includes linear interpolation of the one-hot controls, yielding nonlinear motion of the corresponding synthetic markers. The range of motion (ROM) generally has about 2000 frames of phoneme and dialogue clips. In addition, the artist can add the animations of previously approved shots to enhance the diversity of the training set.

We use Autodesk Maya \cite{maya} to generate the synthetic markers for the training tuples. Artist-created animations are applied to the facial controller attributes to drive the facial puppet. The performance of the generating process is about 5 frames per second with an AMD Ryzen Threadripper PRO 3995WX 64-Core CPU and Quadro RTX A5000 24GB graphics card.

\section{Architecture and Training Setup}
%\section{Training setup}

The FDLS training is performed offline in our production environment. We use the Adam optimizer \cite{kingma:adam} to train the neural networks for each facial region. The region-specific hyper-parameters are given in the following table:

\begin{center} %   Group  &   lr  &  dp   &   bs  &  epoch &  l2  & time
\begin{tabular}{ |p{1.5cm}||p{.8cm}|p{.4cm}|p{.4cm}|p{.5cm}|p{.8cm}|p{.8cm}| }
 \hline
 \multicolumn{7}{|c|}{Region network hyperparameters} \\
 \hline
 Group      & lr   & dp & bs  & \#ep & l2 & time \\
 \hline
 lower-face & 1e-4 & 0.01 & 64 & 450 & 1e-5 & 20m \\
 upper-face & 5e-4 & 0.01 & 64 & 300 & 1e-5 & 15m \\
 cheek      & 1e-4 & 0.01 & 64 & 300 & 1e-5 & 15m \\
 eyelids    & 2e-4 & 0.0  & 64 & 300 & 1e-5 & 10m \\
 eyeball    & 1e-4 & 0.0  & 64 & 250 & 1e-5 & 5m \\
 jaw        & 1e-4 & 0.01 & 64 & 300 & 1e-5 & 15m \\
 lips       & 1e-4 & 0.01 & 64 & 450 & 1e-5 & 20m \\
 \hline
\end{tabular}
\end{center}

%The lr, dp, bs, ep, l2, time denote the learning rate, dropout, batch size, epochs, L2-norm coefficient and training time, respectively.
The learning rate, dropout, batch size, epochs, L2-norm coefficient and training time are denoted as lr, dp, bs, ep, l2, time, respectively.
We highlight that the 3d eye marker set shown in \Figref{img:marker} is constructed by ray-casting from the upper camera center through the 2d tracked eye markers in the dewarped head camera footage. The intersection between the ray and the face puppet determines the 3d eye marker. We find that the projected eye marker set is relatively more stable than other regions, thus we remove dropout in training the eyelids and eyeball.

The next table shows the per-region network architectures:
%The following table shows the neural network configurations for different facial regions:
\begin{center} %   Group  & feature& in dim& our dim& n hidden
\begin{tabular}{ |p{1.5cm}||p{1.5cm}|p{.5cm}|p{.4cm}|p{.8cm}| }
 \hline
 \multicolumn{5}{|c|}{Region network architectures} \\
 \hline
 Group      & feature & rb dim & \#rb & jaw cond?\\
 \hline
 lower-face & dist-delta & 800 & 3 & \checkmark \\
 upper-face & dist-dir   & 300 & 3 & \\
 cheek      & dist-delta & 500 & 2 & \checkmark \\
 eyelids    & dist-dir   & 300 & 2 & \\
 eyeball    & dist-dir   & 150 & 2 & \\
 jaw        & dir-delta  & 800 & 2 & \\
 lips       & dir        & 600 & 3 & \checkmark \\
 \hline
\end{tabular}
\end{center}
All the neural networks are fully-connected ResNets \cite{HeZRS15}, each with an input layer followed with a varied number of residual blocks (\#rb) finally followed by layers to match target dimensions. Furthermore, the jaw kinematic conditioning values are concatenated to the first hidden layer following a series of residual blocks.

\section{Facial Region Definitions}

\textbf{Controls per region.}
The following table shows the FACS controls (i.e.~blendshape weights or channels) for each face region. The face puppet in our paper is constructed with symmetrically split channel configuration, for instance, cheekRaiserR and cheekRaiserL represent the right and left-side cheekRaiser respectively. In the table we list the generic (not split) name for brevity.\\

\begin{figure}[H]
    \centering
    \includegraphics[width=0.45\textwidth]{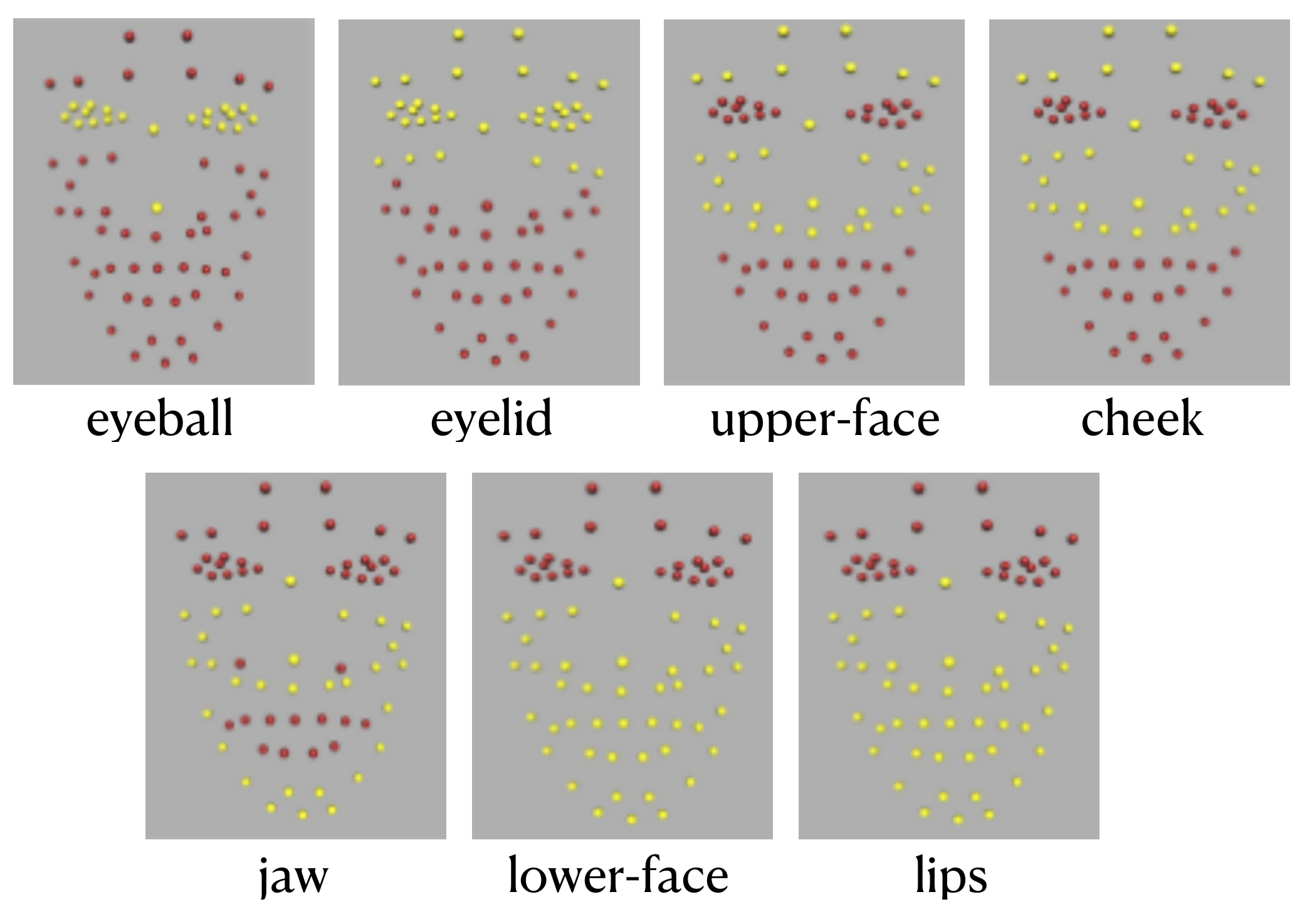}
    %\caption{The marker sets for facial grouping}
    \caption{The per-region marker groups.}
    \label{img:marker}
\end{figure}

\begin{tabular}{ |p{2cm}|p{2.1cm}|p{1.5cm}|p{1.5cm}|  }
 \hline
 \multicolumn{4}{|c|}{FACS controls used per facial region} \\
 \hline
 upper-face      & cheek        & lower-face & lips \\
 \hline
 innerBrowRaiser & cheekRaiser  & incisivus & lipPuckerer \\
 outerBrowRaiser & noseWrinkler & lipRaiser & lipFunneler \\
 procerus        & cheekPuff    & lipDepressor & lipTightener\\
 browLowerer     & noseDepressor & chinRaiser & innerOO     \\
                 & nasolabialFurrow & lipPressor & outerOO     \\
                 & lidTightener     & incisivus & \\
                 & squint & & \\
 \hline
\end{tabular}
\\\\
``OO'' denotes the orbicularis oris muscle.
These remaining regions were omitted from the table due to limited space:
%We show the remaining channel group setting due to the limited table space.
\textit{jaw}: jawOpen, jawSideway, jawThrust. \textit{eyeball}: eyeLeftRight, eyeUpDown. \textit{eyelids}: eyeClose.
\\\\
\textbf{Markers per region.}
\Figref{img:marker} shows the marker selection for each facial region.

% We empirically select the markers from the full marker set for training in the following \Figref{img:marker}.}
Notice that although upper-face/cheek and lower-face/lips have the same marker set, the features of each group are different and thus the neural network training behaves differently. The ``lips'' group especially benefits from the pairwise direction feature as the depth information is particularly important in this region.

% single image

% \bibliographystyle{ACM-Reference-Format}
% \bibliography{reference/references}

\end{document}